\documentclass[letterpaper, 10 pt, conference]{ieeeconf}

\IEEEoverridecommandlockouts
\usepackage{amsmath,amssymb,amsfonts}
\usepackage{mathtools}

\usepackage{amsthm}

\theoremstyle{remark}

\usepackage{array}
\usepackage{booktabs}
\usepackage[table]{xcolor}
\usepackage{multirow}
\usepackage{tabularx}
\usepackage{makecell}
\usepackage{caption}
\usepackage{multicol}
\newcolumntype{C}[1]{>{\centering\arraybackslash}m{#1}}

\usepackage{graphicx}
\usepackage{epsfig}

\usepackage{algorithm}
\usepackage{algpseudocode}

\usepackage{cuted}
\usepackage{capt-of} 
\usepackage[font=footnotesize]{caption}

\usepackage{array,booktabs}
\usepackage[table]{xcolor}
\usepackage{caption}
\usepackage{subcaption}
\newcolumntype{C}[1]{>{\centering\arraybackslash}m{#1}}

\usepackage{pifont}
\newcommand{\cmark}{\ding{51}}
\newcommand{\xmark}{\ding{55}}
\newcommand{\warn}{{\fontencoding{U}\fontfamily{futs}\selectfont\char49\relax}}

\usepackage{cite}

\usepackage{hyperref}
\hypersetup{
    colorlinks=true,
    citecolor=cyan
}
\usepackage{orcidlink}

\title{ \LARGE \bf
Safe and Optimal Variable Impedance Control via Certified Reinforcement Learning}
\author{
Shreyas Kumar$^{\orcidlink{0009-0003-2869-4913}}$ and
Ravi Prakash$^{\orcidlink{0000-0002-9058-434X}}$ \\
\url{shr-eyas.github.io/CGMS}
\thanks{Both authors are with Human-interactive Robotics Lab, IISc Bangalore.}
\thanks{This work was supported in part by ARTPARK, IISc Bangalore.}
\thanks{Corresponding author: {\tt\small shreyaskumar@iisc.ac.in}}
}

\begin{document}
\maketitle
\thispagestyle{empty}
\pagestyle{empty}
\vspace*{-150mm}

\begin{strip}
\centering
\includegraphics[width=0.9\textwidth]{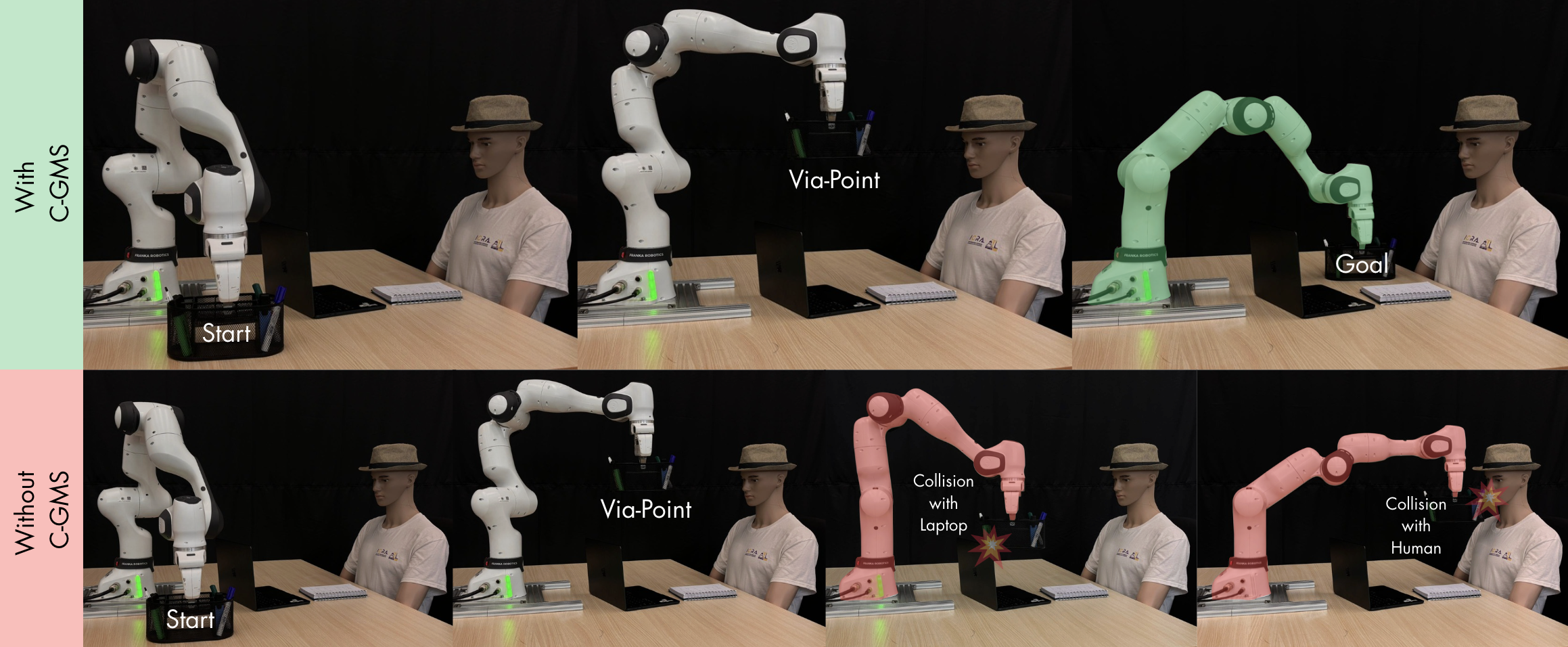}
\captionof{figure}{Comparison of policy execution under certified vs. unconstrained learning. Top: With (proposed) C-GMS, policy sampling is restricted to a certified manifold, ensuring Lyapunov stability and safe execution through a via-point to the goal. Bottom: Without C-GMS, unconstrained sampling may violate the stability conditions, leading to unsafe behaviors, including collisions with the environment/human.}
\label{fig:placeholder}
\end{strip}

\begin{abstract}

Reinforcement learning (RL) offers a powerful approach for robots to learn complex, collaborative skills by combining Dynamic Movement Primitives (DMPs) for motion and Variable Impedance Control (VIC) for compliant interaction. However, this model-free paradigm often risks instability and unsafe exploration due to the time-varying nature of impedance gains. This work introduces Certified Gaussian-Manifold Sampling (C-GMS), a novel trajectory-centric RL framework that learns combined DMP and VIC policies while guaranteeing Lyapunov stability and actuator feasibility by construction. Our approach reframes policy exploration as sampling from a mathematically defined manifold of stable gain schedules. This ensures every policy rollout is guaranteed to be stable and physically realizable, thereby eliminating the need for reward penalties or post-hoc validation. Furthermore, we provide a theoretical guarantee that our approach ensures bounded tracking error even in the presence of bounded model errors and deployment-time uncertainties. We demonstrate the effectiveness of C-GMS in simulation and verify its efficacy on a real robot, paving the way for reliable autonomous interaction in complex environments.
\end{abstract}
\section{INTRODUCTION}

\begin{table*}[!t]
\centering
\captionsetup{font=scriptsize}
\renewcommand{\arraystretch}{1.3}
\begin{tabular}{
| >{\centering\arraybackslash} m{2.75cm}
| >{\centering\arraybackslash} m{2.0cm}
| >{\centering\arraybackslash} m{2.0cm}
| >{\centering\arraybackslash} m{2.5cm}
| >{\centering\arraybackslash} m{1.5cm}
| >{\centering\arraybackslash} m{2.0cm}
| >{\centering\arraybackslash} m{1.5cm}|
}
\hline
    \textbf{Method Family} & 
    \textbf{Stability Aware (VIC)} &
    \textbf{Actuator-Limit Aware} &
     \textbf{Handles Non-linear Dynamics} &
    \textbf{Optimizes Task Cost} &
    \textbf{Adaptive Gains} &
    \textbf{Model-Free Learning}  \\
    \hline

    iLQR / DDP / Trajectory-MPC
    &
    \warn\  (local CLF constraints) &
    \warn\ (include as constraint)  &
    \warn\ (local linearization) &
    \cmark\  &
    \warn\ (pre-schedule) &
    \xmark  \\
    \hline

    CBF / RMP Safety Filters
    &
    \warn\ (safety invariance, not VIC-specific) &
    \warn\ (included as constraint) &
    \cmark  &
    \xmark &
    -- &
    --  \\
    \hline

    Energy-Tank / PO-PC
    &
    \cmark &
    \xmark\ (passivity/power aware) &
    \cmark &
    \xmark &
    \cmark &
    --  \\
    \hline

    Model-based VIC learning
    & 
    \xmark &
    \xmark &
    \cmark &
    \cmark &
    \cmark &
    \xmark  \\
    \hline
    
    PI$^2$ / PI-BB VIC
    &
    \xmark &
    \xmark &
    \cmark &
    \cmark &
    \cmark &
    \cmark \\
    \hline

    \textbf{Our method (PI$^2$ + C-GMS)} &
    \cmark &
    \cmark &
    \cmark &
    \cmark &
    \cmark &
    \cmark \\
    \hline
  \end{tabular}
  \caption{Comparison of method families for learning variable impedance control policies. Our method is the only one to combine model-free learning, gain scheduling, stability certification, and actuator feasibility within a unified optimization framework.}
  \label{tab:comparison_table}
  \vspace{-15pt}
\end{table*}

The field of robotics is undergoing a fundamental shift, moving from static, repetitive industrial tasks to dynamic, unstructured environments where physical interaction is not only inevitable but essential. To navigate this new paradigm, robots require the ability to adapt learned behaviors to new settings, which is often achieved through task parameterization using representations like Dynamic Movement Primitives (DMPs) \cite{saveriano2021dynamic}. While DMPs excel at generalizing a movement to new endpoints, handling complex motions that require passing through a series of via-points is not trivial. For this, and for safe physical interaction, robots also need the ability to dynamically adjust their stiffness and damping, a capability provided by Variable Impedance Control (VIC).

The complexity of designing time-varying impedance profiles has motivated the adoption of data-driven methods, with Reinforcement Learning (RL) emerging as a powerful approach. In model-based families, such as Differential Dynamic Programming (DDP) \cite{mayne1973differential}, Iterative Linear Quadratic Regulator (iLQR) \cite{li2004iterative}, and Model Predictive Control (MPC) \cite{rawlings2020model}, a robot’s dynamics model is exploited to compute optimal sequences. These methods can incorporate actuator and stability constraints via Control Lyapunov Functions (CLFs), but they are computationally expensive, depend heavily on model accuracy, and typically pre-schedule gains rather than learning them adaptively.

Model-free approaches avoid explicit dynamics modeling, enabling the discovery of complex behaviors in unstructured environments. RL, particularly Path Integral (PI$^2$) \cite{kappen2005linear,theodorou2010generalized}, has proven effective for learning intricate motor skills. Seminal work by Buchli et al. \cite{buchli2011variable} pioneered the use of PI$^2$ to simultaneously learn DMP trajectory parameters and time-varying impedance gains, laying the groundwork for data-driven compliant control. Later, Rey et al. \cite{rey2018learning} refined this approach by incorporating human demonstrations, improving efficiency. While highly successful at optimizing task costs, these methods, like many in the safe RL literature that use reward penalties \cite{achiam2017constrained} or post-hoc safety filters \cite{CBFAmes}, do not explicitly enforce stability during learning, leaving Variable Impedance Control susceptible to instability, as rigorously proven by Kronander and Billard \cite{KBVIC}.

Alternative approaches have focused on safety through other principles. Passivity-based controllers, such as Energy-Tank methods \cite{passivityHaptic,tankVIC,impedanceEnergyTank}, guarantee stability by maintaining passivity, but they cannot directly optimize general task costs. Safety Filters, such as Control Barrier Functions (CBFs) \cite{CBFAmes} and Reactive Motion Planning (RMP) \cite{cheng2019rmpflowcomputationalgraphautomatic}, provide generic safety invariance layers, but they are not tailored to the time-varying nature of VIC and are not tightly integrated into policy search. Model-based VIC learning approaches \cite{anand2023model} optimize gains directly but remain limited by modeling assumptions and lack guarantees during exploration. As summarized in Table \ref{tab:comparison_table}, current methods lack built-in stability and actuator-limit awareness, posing a significant risk of instability in VIC~\cite{KBVIC}.

In this work, we introduce Certified Gaussian-Manifold Sampling (C-GMS), a novel reinforcement learning framework that unifies model-free policy search with rigorous stability analysis. Unlike prior methods that penalize instability or apply safety filters post hoc, our approach guarantees Lyapunov stability \cite{KBVIC} and actuator feasibility by construction. By embedding the stability criterion directly into the RL exploration loop, we restrict policy sampling to a certified manifold of stable gain schedules. Each Gaussian perturbation is analytically sampled from this manifold, ensuring that every single rollout remains stable and physically realizable. This integration eliminates the need for separate safety critics or penalty terms and enables reliable policy optimization in complex tasks. Furthermore, we establish a formal theorem demonstrating that our method not only ensures internal stability but also guarantees uniform ultimate boundedness of the tracking error in presence of model and sensor inaccuracies, thus providing a strong foundation for its applicability in real-world scenarios.


\section{PRELIMINARIES}
\label{sec:prelim}
\subsection{Rigid Body Dynamics}
We assume that the joint-space rigid-body dynamics of an $n$-DoF manipulator is given by $\mathbf{M}(\mathbf{q})\,\ddot{\mathbf{q}} + \mathbf{C}(\mathbf{q},\dot{\mathbf{q}})\,\dot{\mathbf{q}} + \mathbf{g}(\mathbf{q}) \;=\; \boldsymbol{\tau}_c + \boldsymbol{\tau}_e,$
where $\mathbf{q},\dot{\mathbf{q}},\ddot{\mathbf{q}} \in \mathbb{R}^{n}$ denote joint position, velocity, and acceleration, $\mathbf{M}(\mathbf{q})\!\succ\!\mathbf{0}$ the inertia matrix, $\mathbf{C}(\mathbf{q},\dot{\mathbf{q}})\,\dot{\mathbf{q}}$ the Coriolis/centrifugal term, $\mathbf{g}(\mathbf{q})$ the torque due to gravity, and $\boldsymbol{\tau}_c,\boldsymbol{\tau}_e \in \mathbb{R}^{n}$ are the commanded and external torques. In task space with Jacobian $\mathbf{J}(\mathbf{q})\in\mathbb{R}^{m\times n}$, the operational-space inertia~\cite{khatib2003unified} is defined as $\boldsymbol{\Lambda}(\mathbf{q}) \;=\; (\mathbf{J}(\mathbf{q})\,\mathbf{M}^{-1}(\mathbf{q})\,\mathbf{J}^\top(\mathbf{q}))^{-1}  \in \mathbb{R}^{m\times m}$. Let $\boldsymbol{\mu}(\mathbf{q},\dot{\mathbf{q}})\!\in\!\mathbb{R}^{m}$ collect Coriolis/centrifugal wrenches and $\mathbf{p}(\mathbf{q})\!\in\!\mathbb{R}^{m}$ be the gravity wrench mapped to task space. Then the end-effector dynamics reads
\begin{equation}
    \boldsymbol{\Lambda}(\mathbf{q})\,\ddot{\mathbf{x}} + \boldsymbol{\mu}(\mathbf{q},\dot{\mathbf{q}}) + \mathbf{p}(\mathbf{q}) \;=\; \mathbf{f}_c + \mathbf{f}_e,
    \label{eq:os-plant}
\end{equation}
where $\mathbf{x},\dot{\mathbf{x}},\ddot{\mathbf{x}} \in \mathbb{R}^{m}$ are the task-space position, velocity, and acceleration, and $\mathbf{f}_c,\mathbf{f}_e\!\in\!\mathbb{R}^{m}$ are the commanded and external wrenches. 
Torques are obtained via the wrench-torque map
$\boldsymbol{\tau}_c = \mathbf{J}^\top(\mathbf{q})\, \mathbf{f}_c.$
These relations follow the standard operational-space formulation and fix the notation used throughout.

\subsection{Variable Impedance Control}
Let $\tilde{\mathbf{x}} = \mathbf{x} - \mathbf{x}_d$ and $\dot{\tilde{\mathbf{x}}} = \dot{\mathbf{x}} - \dot{\mathbf{x}}_d$ denote the Cartesian error and its velocity, where the reference $\mathbf{x}_d(t)$ is assumed to be twice differentiable. 
We shape the interaction behavior using time-varying symmetric positive-definite gain schedules $\mathbf{K}(t)=\mathbf{K}^\top(t)\succ\mathbf{0}$, and $\mathbf{D}(t)=\mathbf{D}^\top(t)\succ\mathbf{0}$
The desired task-space inertia is fixed to a constant matrix $\mathbf{H} = \mathbf{H}^\top \succ \mathbf{0}$.
Under operational-space inverse-dynamics (OSID), we command the wrench $\mathbf{f}_c$
\begin{equation}
\mathbf{f}_c = \boldsymbol{\Lambda} \ddot{\mathbf{x}}_{\text{cmd}} 
+ \boldsymbol{\mu} 
+ \mathbf{p} 
+ \underbrace{\left( \boldsymbol{\Lambda} \mathbf{H}^{-1} - \mathbf{I} \right) \mathbf{f}_e}_{\text{feedforward term} \,\mathbf f_\mathrm{f}},
\label{eq:fc}
\end{equation}
where the commanded acceleration is defined as 
\begin{equation}
\ddot{\mathbf{x}}_{\mathrm{cmd}}
\, = \,
\ddot{\mathbf{x}}_d - \mathbf{H}^{-1} \, \big(\,\mathbf{D}(t) \, \dot{\tilde{\mathbf{x}}} + \mathbf{K}(t) \, \tilde{\mathbf{x}} \,\big).
  \label{eq:xdd_cmd}
\end{equation}
Substituting \eqref{eq:fc}, \eqref{eq:xdd_cmd} into operational-space dynamics \eqref{eq:os-plant} and simplifying, the closed-loop error dynamics become
\begin{equation}
\mathbf{H} \, \ddot{\tilde{\mathbf{x}}} + \mathbf{D}(t) \, \dot{\tilde{\mathbf{x}}}
+ \mathbf{K}(t) \, \tilde{\mathbf{x}}
\, = \, \mathbf{f}_e.
\label{eq:closed_loop}
\end{equation}
This matches the classical form of a time-varying impedance behavior with desired inertia $\mathbf{H}$.

\subsection{Stability in VIC}
Under dynamic decoupling with a constant desired inertia $\mathbf{H}\succ\mathbf{0}$ and in free space ($\mathbf{f}_e=\mathbf{0}$), Kronander \& Billard~\cite{KBVIC} show that the closed-loop impedance dynamics in~\eqref{eq:closed_loop} are globally uniformly stable if there exists $\alpha>0$ such that, for all $t$,
\begin{align}
  &\alpha\,\mathbf{H} - \mathbf{D}(t) \;\preccurlyeq\; \mathbf{0} \quad  \text{and,} \nonumber \\ 
  &\dot{\mathbf{K}}(t) + \alpha\,\dot{\mathbf{D}}(t) - 2\alpha\,\mathbf{K}(t) \;\preccurlyeq\; \mathbf{0}.
  \label{eq:KB}
\end{align}
These are state-independent, pointwise-in-time constraints on the gain schedules that can be constructed offline. In our setting, the external wrench $\mathbf{f}_e$ appears explicitly on the right-hand side of the dynamics, while Eq.~\eqref{eq:KB} certifies the stability of the internal (unforced) system. This ensures that the impedance behavior defined by $(\mathbf{H},\,\mathbf{D}(t),\,\mathbf{K}(t))$ is a dissipative and well-posed map throughout the motion.

\subsection{Policy Improvement with Path Integrals}
\label{subsec:prelim_pi2}

We consider the problem of minimizing the expected trajectory cost
\[
J(\tau) = \Phi\big(\mathbf{x}(T)\big) + \int_{0}^{T} \ell\big(\mathbf{x}(t), \mathbf{u}(t), t\big)\, dt ,
\]
where \(\mathbf{x}(t) \in \mathbb{R}^{n_x}\) is the system state, \(\mathbf{u}(t) \in \mathbb{R}^{n_u}\) is the control input, \(\ell(\cdot)\) is the running cost, and \(\Phi(\cdot)\) is the terminal cost. The dynamics are assumed to be control-affine with additive noise:
\[
\dot{\mathbf{x}} = \mathbf{f}(\mathbf{x},t) + \mathbf{G}(\mathbf{x},t)\,\mathbf{u}(t) + \mathbf{G}(\mathbf{x},t)\,\boldsymbol{\xi}(t),
\]
where \(\boldsymbol{\xi}(t)\) is zero-mean Gaussian noise. When the noise covariance and control penalty satisfy \(\boldsymbol{\Sigma}_\xi = \lambda\, \mathbf{R}^{-1}\), a logarithmic transformation of the value function yields a linear HJB equation. Following established results in stochastic optimal control~\cite{kappen2005linear, theodorou2010generalized}, the solution admits the form
\[
\Psi(\mathbf{x}_i, t_i) = \mathbb{E} \left[\exp\left( -\tfrac{1}{\lambda} \left( \Phi(\mathbf{x}_T) + \int_{t_i}^{T} q(\mathbf{x}(t), t)\, dt \right) \right) \right],
\]
where \(q(\cdot)\) is the state cost, and the expectation is over stochastic trajectories initiated at \((\mathbf{x}_i, t_i)\).

Thus the stochastic optimal control problem becomes a path-integral estimation problem. Rather than computing the value function explicitly, the optimal control can be written as an expectation over trajectories \cite{kappen2005linear,theodorou2010generalized}:
\begin{equation}
\mathbf{u}_{t_i} \;=\; \int P(\tau_i)\,\mathbf{u}(\tau_i)\, d\tau_i ,
\label{eq:pi2_control}
\end{equation}
with \(\mathbf{u}(\tau_i) = \mathbf{R}^{-1}\mathbf{G}^{\top} \big(\mathbf{G}\mathbf{R}^{-1}\mathbf{G}^{\top}\big)^{-1} \big(\mathbf{G}\,\boldsymbol{\xi}_{t_i} - \mathbf{b}_{t_i}\big)\).

Here \(P(\tau_i)\) is the probability density of a trajectory segment \(\tau_i\) starting at \((\mathbf{x}_i,t_i)\), and \(\mathbf{b}_{t_i}\) collects drift and cost terms (see \cite{theodorou2010generalized} for the explicit expression). PI$^2$ evaluates Eq.~\eqref{eq:pi2_control} by Monte Carlo rollouts, i.e., it approximates the path-integral expectation with a sample-weighted average over trajectories. (cf. Table 1 in~\cite{buchli2011variable} for the algorithm).

\section{METHOD}
\label{sec:method}

\begin{figure*}[h]
    \centering
    \includegraphics[width=1.0\linewidth]{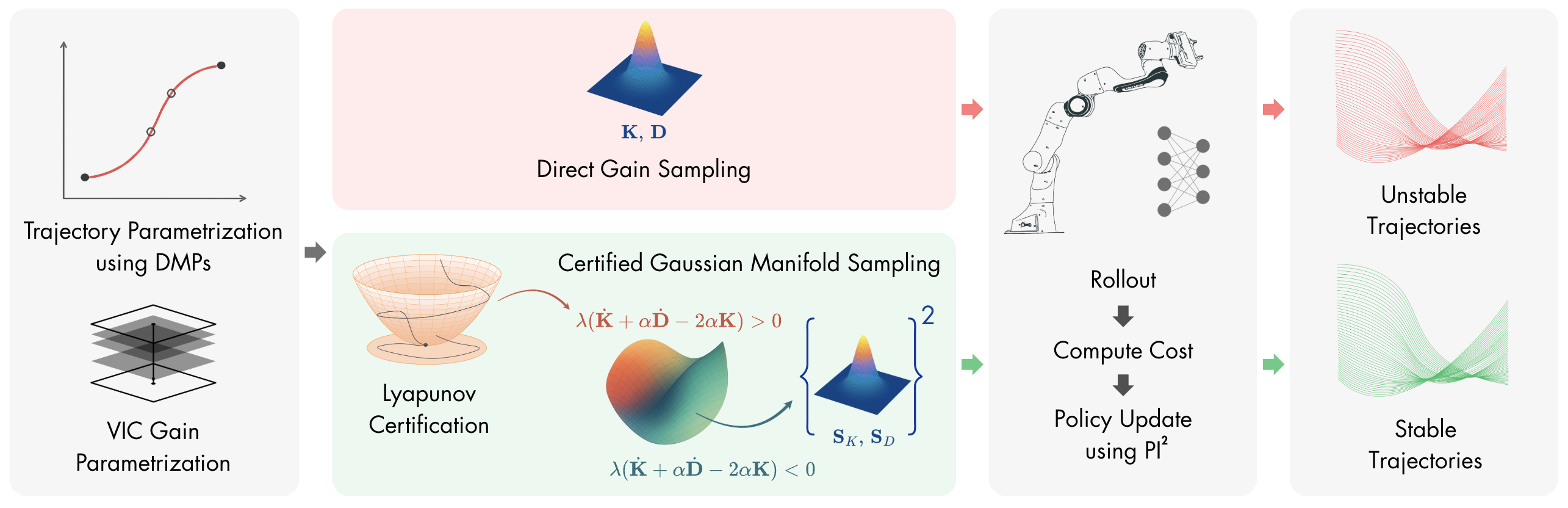}
    \caption{Overview of the C-GMS framework. Trajectories are parameterized using DMPs, and time-varying VIC gains are parameterized using slacks. In a standard approach, gains are sampled directly from a Gaussian, which can violate Lyapunov stability conditions and lead to unstable rollouts. In contrast, C-GMS enforces stability by sampling from a certified manifold where the Lyapunov condition~\cite{KBVIC} holds resulting in stable trajectories throughout learning.} 
    \label{fig:placeholder}
    \vspace{-15pt}
\end{figure*}

We present \emph{Certified Gaussian-Manifold Sampling} (C-GMS), a trajectory-centric reinforcement learning framework for variable impedance control (VIC) that guarantees Lyapunov stability and actuator feasibility during exploration. C-GMS samples policies via Gaussian perturbations in parameter space, but each sample is from a manifold where safety and stability conditions are enforced analytically. This manifold is defined by a time-varying Lyapunov certificate (cf. Eq.~\eqref{eq:KB}), and gain schedules are synthesized using slack variables that satisfy these conditions by construction. As a result, every rollout, regardless of the sampled policy, remains certified--yielding stable, safe, and physically realizable interaction. C-GMS thus combines model-free policy learning with model-based guarantees, eliminating the need for penalty terms, barriers, or post-hoc projection.

\subsection{Trajectory Parametrization via DMPs}
PI$^2$ computes the optimal control update (cf. Eq.~\eqref{eq:pi2_control}) for a system with a parameterized policy $\mathbf{a}_t = \boldsymbol{\Phi}_t\,(\boldsymbol{\theta} + \boldsymbol{\xi}_t)$, where \(\boldsymbol{\theta}\) is a learned parameter vector and \(\boldsymbol{\xi}_t\) is exploration noise. In C-GMS, Gaussian perturbations are sampled once per episode, \(\boldsymbol{\xi} \sim \mathcal{N}(\mathbf{0}, \boldsymbol{\Sigma})\), and applied consistently across time, i.e., \(\boldsymbol{\xi}_t = \boldsymbol{\xi}\) for all \(t\). The basis functions are defined by
\[
[\Phi(s_t)]_j = \frac{\Psi_j(s_t)}{\sum_{k=1}^M \Psi_k(s_t)}, \quad
\Psi_j(s_t) = \exp\!\left(-\tfrac{(s_t - c_j)^2}{2\sigma_j^2}\right),
\]
with canonical phase \(s_t = 1 - t/\tau\), where \(M\) is the number of radial basis functions (RBFs).

To ensure smooth, structured control signals compatible with PI$^2$, we parametrize the desired trajectory using Dynamic Movement Primitives (DMPs). Let \((\mathbf{x}(t), \dot{\mathbf{x}}(t), \ddot{\mathbf{x}}(t))\) denote the kinematic state in \(\mathbb{R}^D\) over the horizon \(t \in [0, T]\). The DMP transformation dynamics are defined as
\[
\tau^2 m \ddot{\mathbf{x}}(t) = k(\mathbf{g}(t) - \mathbf{x}(t)) - \tau d \dot{\mathbf{x}}(t) + \gamma(t) f_{\text{forcing}}(t),
\]
where \(\tau\) is the temporal scaling factor, and \(k\), \(d\), \(m\) are the stiffness, damping, and mass parameters, respectively. The goal trajectory \(\mathbf{g}(t)\) may be constant or time-varying, and \(\gamma(t)\) is a phase-dependent scaling term. The nonlinear forcing term \(f_{\text{forcing}}(t)\) modulates the trajectory to encode complex motion patterns, and is modeled as
\[
f_{\text{forcing}}(t) = \Phi_{\mathrm{traj}}(s_t)(\boldsymbol{\theta}_{\mathrm{traj}} + \boldsymbol{\xi}_{\mathrm{traj}}),
\]
where $\Phi_{\mathrm{traj}}(s_t)$ is the normalized RBF vector, and $\boldsymbol{\theta}_{\mathrm{traj}} \in \mathbb{R}^{M \times D}$ parametrizes the forcing profile and thus the trajectory. We next describe how C-GMS ensures that all sampled gain schedules \((\mathbf{D}(t), \mathbf{K}(t))\) satisfy stability and torque feasibility throughout the policy search process.

\subsection{Gain Parametrization via C-GMS}
To extend the PI$^2$ parametrization beyond trajectories, we introduce an analogous representation for time-varying impedance gains. The objective is to ensure that exploration remains confined to a certified manifold, where every sample satisfies Lyapunov stability and feasibility conditions by construction. Following the conditions for global uniform stability in \cite{KBVIC}, we enforce the inequalities structurally by introducing matrix-valued slack variables $\mathbf{S}_D(t)$ and $\mathbf{S}_K(t)$:
\begin{align}
\alpha\,\mathbf{H} - \mathbf{D}(t) &= -\mathbf{S}_D(t)\,\mathbf{S}_D^\top(t) \,\preccurlyeq 0, \label{eq:D_from_slack} \\
\dot{\mathbf{K}}(t) + \alpha\,\dot{\mathbf{D}}(t) - 2\alpha\,\mathbf{K}(t) 
&= -\mathbf{S}_K(t)\,\mathbf{S}_K^\top(t) \,\preccurlyeq 0. 
\end{align}
These slacks are parametrized analogously to the trajectory forcing term using time-varying basis functions:
\begin{align}
\operatorname{vec}_\triangle(\mathbf{S}_D(t)) &= \Phi_{\mathrm{D}}(s_t)\,(\boldsymbol{\theta}_{\mathrm{D}} + \boldsymbol{\xi}_{\mathrm{K}}), \nonumber \\
\operatorname{vec}_\triangle(\mathbf{S}_K(t)) &= \Phi_{\mathrm{K}}(s_t)\,(\boldsymbol{\theta}_{\mathrm{K}} + \boldsymbol{\xi}_{\mathrm{D}}),
\end{align}
where $\Phi_{\mathrm{D}}(s_t)$ and $\Phi_{\mathrm{K}}(s_t)$ share the normalized RBF structure, and the parameter vectors $\boldsymbol{\theta}_{\mathrm{D}}, \boldsymbol{\theta}_{\mathrm{K}} \in \mathbb{R}^{M \times D}$ define the gain profiles.

To ensure $\mathbf{K}(t)$ remains symmetric positive definite (SPD), we evolve its Cholesky factor $\mathbf{Q}(t)$:
\begin{align}
\mathbf{B}(t) &:= -\alpha \dot{\mathbf{D}}(t) - \mathbf{S}_K(t)\,\mathbf{S}_K^\top(t), \nonumber \\
\dot{\mathbf{Q}}(t) &= \alpha\,\mathbf{Q}(t) + \tfrac{1}{2}\,\mathbf{Q}^{-\top}(t)\,\mathbf{B}(t), \nonumber \\
\mathbf{K}(t) &= \mathbf{Q}^\top(t)\,\mathbf{Q}(t),
\label{eq:cholesky-flow}
\end{align}
with $\mathbf{Q}(0)$ initialized such that $\mathbf{K}(0)\succ 0$. Differentiating $\mathbf{K} = \mathbf{Q}^\top \mathbf{Q}$ and substituting \eqref{eq:cholesky-flow} yields
\(\dot{\mathbf{K}}(t) = 2\alpha\,\mathbf{K}(t) + \mathbf{B}(t)\), ensuring that 
\(\dot{\mathbf{K}} + \alpha \dot{\mathbf{D}} - 2\alpha \mathbf{K} = -\mathbf{S}_K \mathbf{S}_K^\top \preccurlyeq 0\) holds identically while maintaining $\mathbf{K}(t)\succ 0$ for all $t$.

This construction confines policy search to a certified manifold of gain schedules that provably satisfy the Lyapunov conditions, eliminating the need for penalties, constraints, or post-hoc projections.



\subsection{Certificate-Aware Gain Contraction}
\label{subsec:cags}
Leveraging C-GMS’s slack-based gain synthesis, which admits a certificate preserving contraction, we impose actuator-limit awareness directly in gain space--without auxiliary constraints, projections, or loss of stability.
This is achieved by introducing a torque governor that uniformly scales the slack variables by $\sqrt{\beta} \in [0,1]$:
$
\mathbf{S}_D^\beta(t) = \sqrt{\beta}\,\mathbf{S}_D(t), \,
\mathbf{S}_K^\beta(t) = \sqrt{\beta}\,\mathbf{S}_K(t),
$
resulting in scaled gains
\begin{align}
 &\mathbf{D}^\beta(t)= \alpha\,\mathbf{H} + \beta\,\mathbf{S}_D(t)\,\mathbf{S}_D^\top(t), \nonumber \\
&\dot{\mathbf{K}}^\beta(t) + \alpha\,\dot{\mathbf{D}}^\beta(t) - 2\alpha\,\mathbf{K}^\beta(t) = -\beta\,\mathbf{S}_K(t)\,\mathbf{S}_K^\top(t).
\end{align}
The gains $\mathbf{K}^\beta(t)$ and $\mathbf{D}^\beta(t)$ remain certificate-compliant for all $\beta \in [0,\,1]$, as they preserve the structure required by the Lyapunov inequalities. Since the control law is affine in both $\mathbf{K}$ and $\mathbf{D}$, the overall torque command is also affine in $\beta$. Denoting the $\beta$-independent component as $\boldsymbol{\tau}_0$ and the $\beta$-dependent component as $\boldsymbol{\tau}_1$, the control input takes the form:
\[
\boldsymbol{\tau}(\beta) = \boldsymbol{\tau}_0 + \beta\,\boldsymbol{\tau}_1,
\]
the maximum admissible scaling factor $\beta^\star$ under actuator box constraints $\boldsymbol{\tau}_{\min} \leq \boldsymbol{\tau}(\beta) \leq \boldsymbol{\tau}_{\max}$ is
\[
\beta^\star = \min_i
\begin{cases}
\frac{\tau_{\max,i} - \tau_{0,i}}{\tau_{1,i}}, & \tau_{1,i} > 0, \\[6pt]
\frac{\tau_{\min,i} - \tau_{0,i}}{\tau_{1,i}}, & \tau_{1,i} < 0,
\end{cases} \quad \in [0,1].
\]
The controller is executed with $\mathbf{S}_{(\cdot)}^{\beta^\star}$ and corresponding gains. This actuator-limit governor preserves the Lyapunov certificate pointwise, enabling safe execution under torque saturation.

\subsection{Convergence Guarantee}
\label{subsec:conv}
We analyze the closed-loop stability of the VIC under time-varying gains synthesized via the C-GMS framework. Specifically, we aim to show that the tracking error remains bounded, even in the presence of model mismatch and other deployment-time uncertainties.

At each time step $t$, the desired trajectory $(\mathbf{x}_d(t),\,\dot{\mathbf{x}}_d(t),\,\ddot{\mathbf{x}}_d(t))$ is generated by evaluating the DMP forcing function $f_{\mathrm{forcing}}(t) = \Phi_{\mathrm{traj}}(s_t)\,\boldsymbol{\theta}_{\mathrm{traj}}$, followed by integration of the DMP dynamics. Simultaneously, the time-varying gains are synthesized via basis expansions $\Phi_D(s_t),\,\Phi_K(s_t)$, which define slack variables $\mathbf{S}_D^{\{.\}}(t), \mathbf{S}_K^{\{.\}}(t)$. These are converted into damping and stiffness gains $\mathbf{D}(t),\,\mathbf{K}(t)$ using equations~\eqref{eq:D_from_slack} and~\eqref{eq:cholesky-flow}, and used in the OSID control law~\eqref{eq:fc}-\eqref{eq:xdd_cmd}.

The deployment of a trained policy in real-world settings brings along challenges such as plant-model mismatch, feedforward wrench $\mathbf{f}_{\mathrm{f}}$ errors, and sensor noise, all of which may induce a residual input $\mathbf{u}_{\mathrm{res}}(t)$ into the closed-loop system. To this end, we establish that the gains generated by C-GMS ensure a bounded tracking error under these conditions.

\noindent \textbf{Theorem.}
Consider the perturbed dynamics under VIC with model error and feedforward uncertainty aggregated as a bounded residual input $\|\mathbf{u_{\mathrm{res}}}(t)\| \le \bar{u} < \infty$. If the gains $\mathbf{D}(t),\,\mathbf{K}(t)$ satisfy the differential Lyapunov conditions~\eqref{eq:A-Vdot-raw} with strict margins $(\varepsilon_D,\,\varepsilon_K) > 0$, then the tracking error $(\tilde{\mathbf{x}}(t),\,\dot{\tilde{\mathbf{x}}}(t))$ is uniformly ultimately bounded, with an ultimate error radius $O(\bar{u}_\mathrm{res})$ that decreases as $(\varepsilon_D,\,\varepsilon_K)$ increase. Proof in Appendix A. \hfill\(\square\)

\section{EXPERIMENTS}
\label{sec:exp}

\subsection{Experimental Setup}
We consider a collaborative human-robot handover task in which a $7$-DoF Franka Research $3$ (FR$3$) manipulator transfers a stationery organizer to a seated human participant. The motivation for the task arises from an everyday scenario: the human is engaged with a notebook and requires a pen, prompting the robot to initiate a handover. The motion is executed under real-time VIC at $1$\,kHz in the task space, with feedback from the robot’s internal force-torque sensors.

\begin{figure}[h]
  \centering
  \begin{subfigure}{\linewidth}
    \centering
    \includegraphics[width=\textwidth]{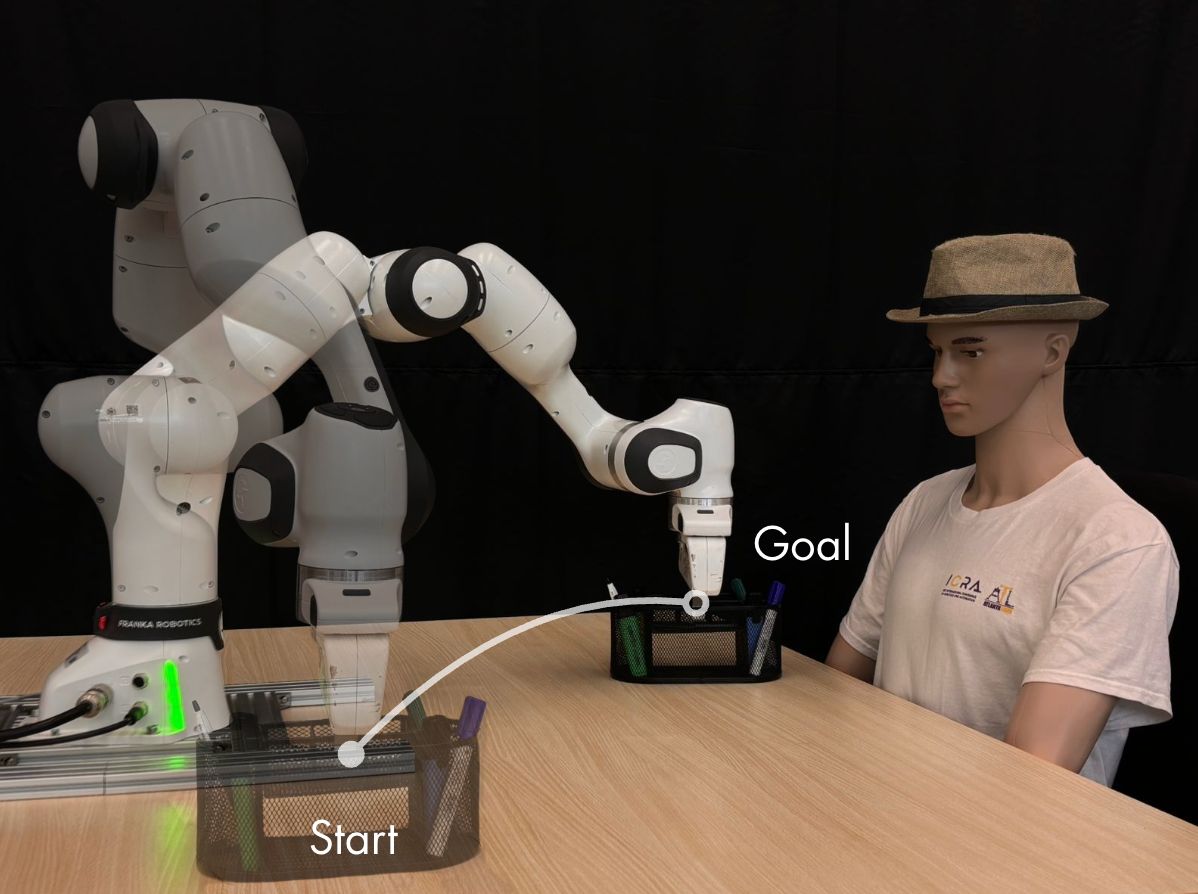}
    \caption{\footnotesize The robot learns to move from a start pose to a predefined goal near the human’s hand, initially following a minimum-jerk trajectory.}
    \label{fig:initial_hw}
  \end{subfigure}

  \vspace{0.2cm}

  \begin{subfigure}{0.48\linewidth}
    \centering
    \includegraphics[width=\textwidth]{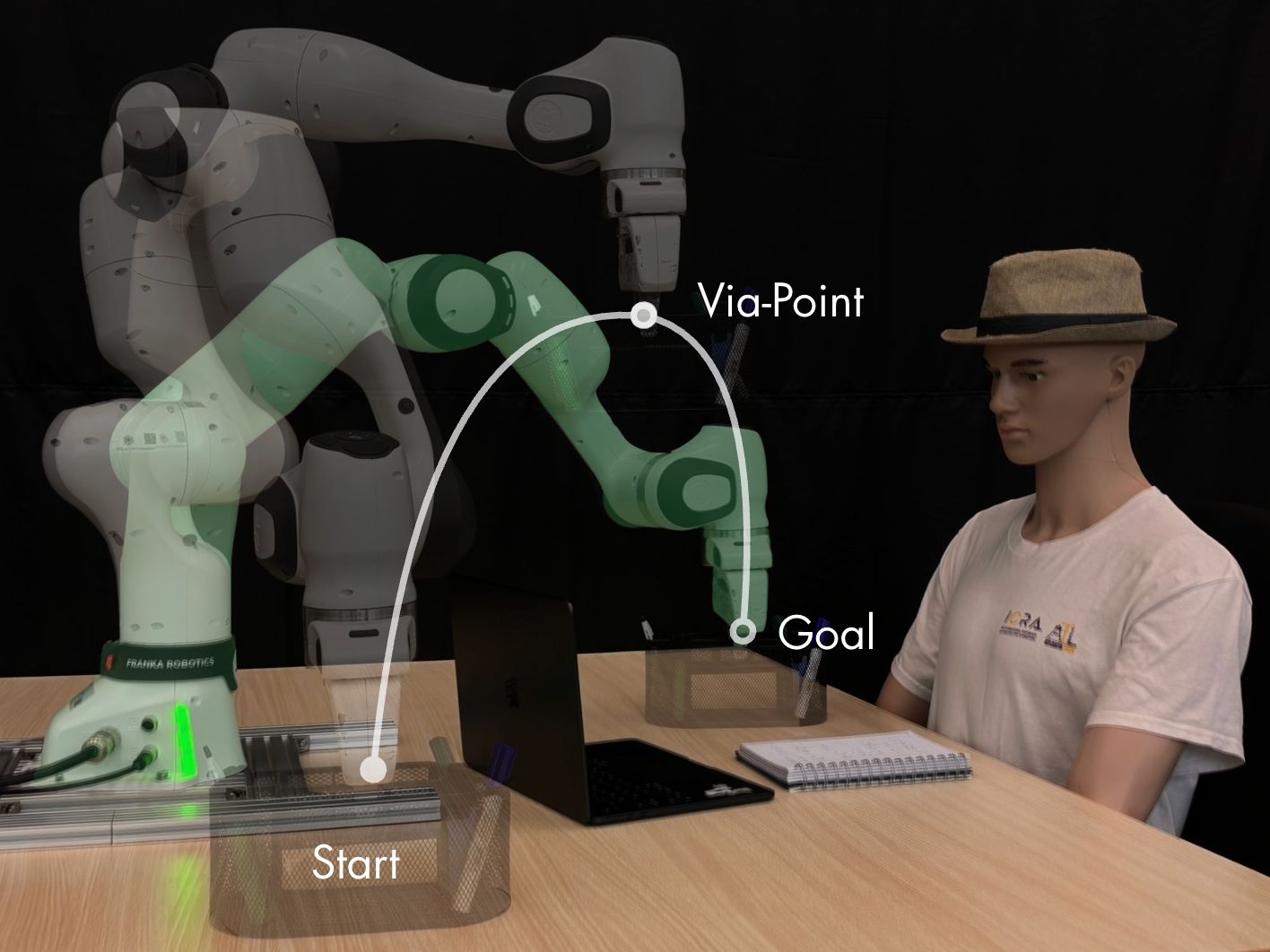}
    \caption{\footnotesize A learned policy under C-GMS introduces a via-point to avoid the obstacle while ensuring stable, compliant behavior resulting in a collision-free trajectory.}
    \label{fig:safe_hw}
  \end{subfigure}
  \hfill
  \begin{subfigure}{0.48\linewidth}
    \centering
    \includegraphics[width=\textwidth]{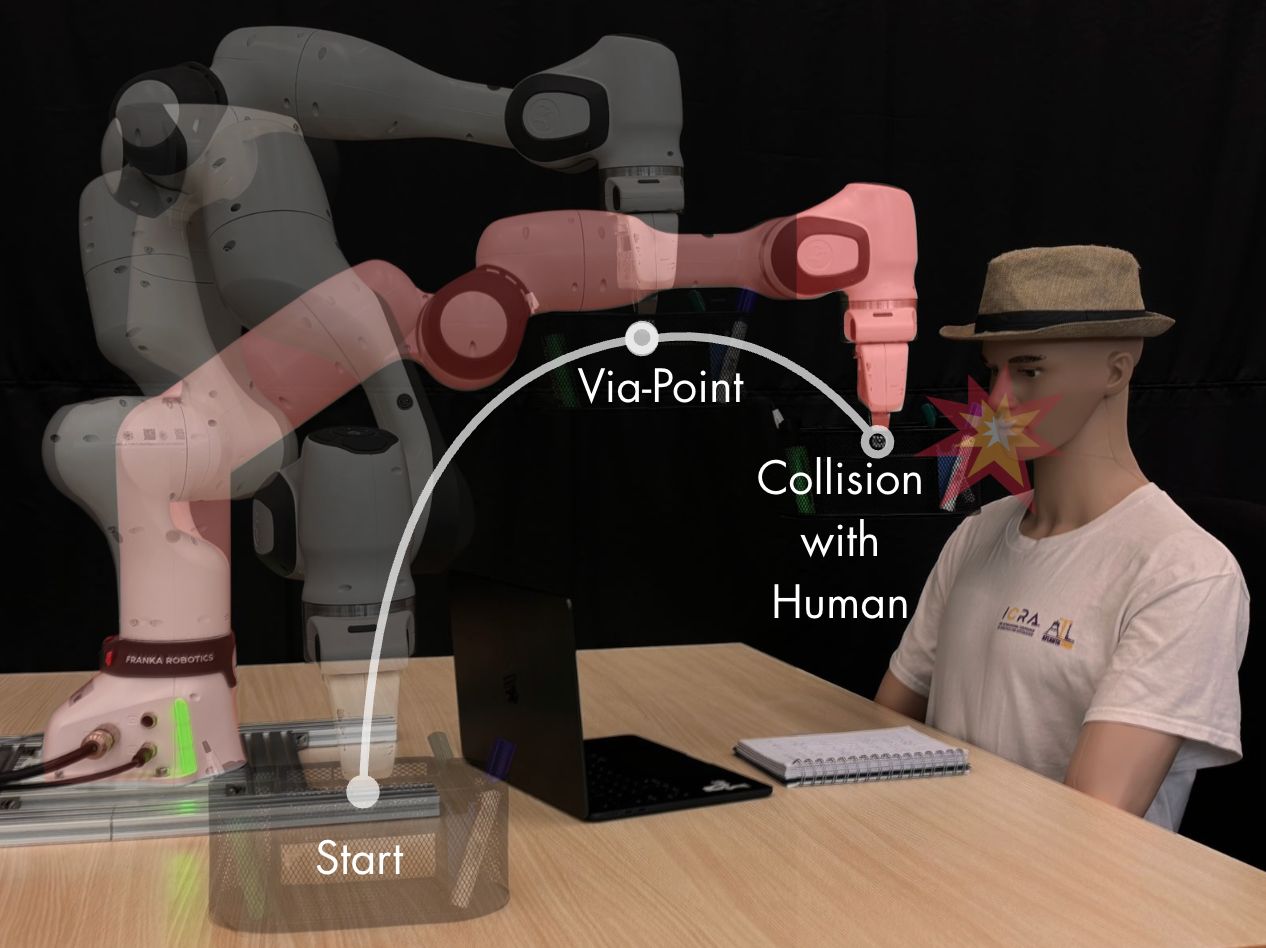}
    \caption{\footnotesize A policy learned without C-GMS violates the Lyapunov condition, resulting in unsafe trajectories and potential collision with the environment or human.}
    \label{fig:unsafe_hw}
  \end{subfigure}
  \caption{Experimental setup for the human-robot collaborative task.}
  \label{fig:hardware}
  \vspace{-15pt}
\end{figure}

\begin{figure*}
\centering
  \begin{subfigure}{0.9\linewidth}
    \centering
    \includegraphics[width=\textwidth]{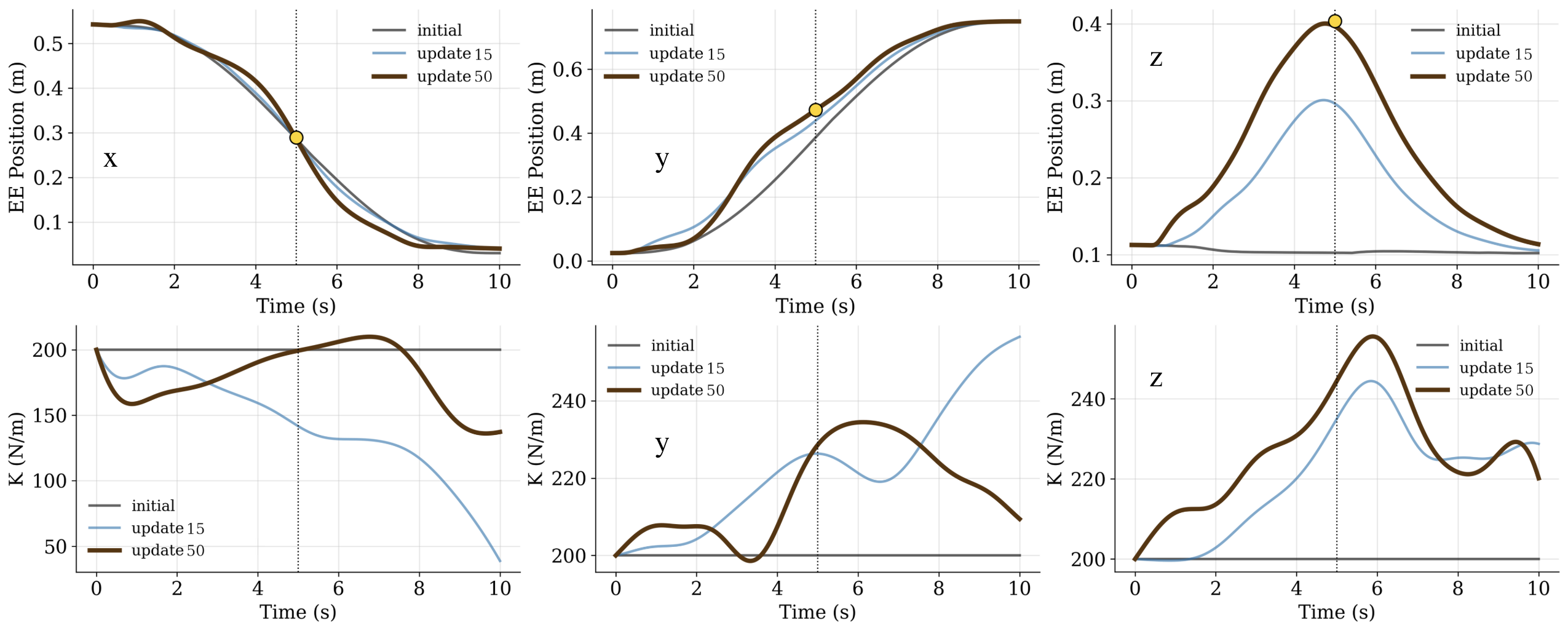}
    \caption{\footnotesize Under full C-GMS, policies remain stable and smooth throughout learning, with impedance gains adapting locally around the via-point}
    \label{fig:safe_res}
  \end{subfigure}
  
    \vspace{0.2cm}
    
  \begin{subfigure}{0.9\linewidth}
    \centering
    \includegraphics[width=\textwidth]{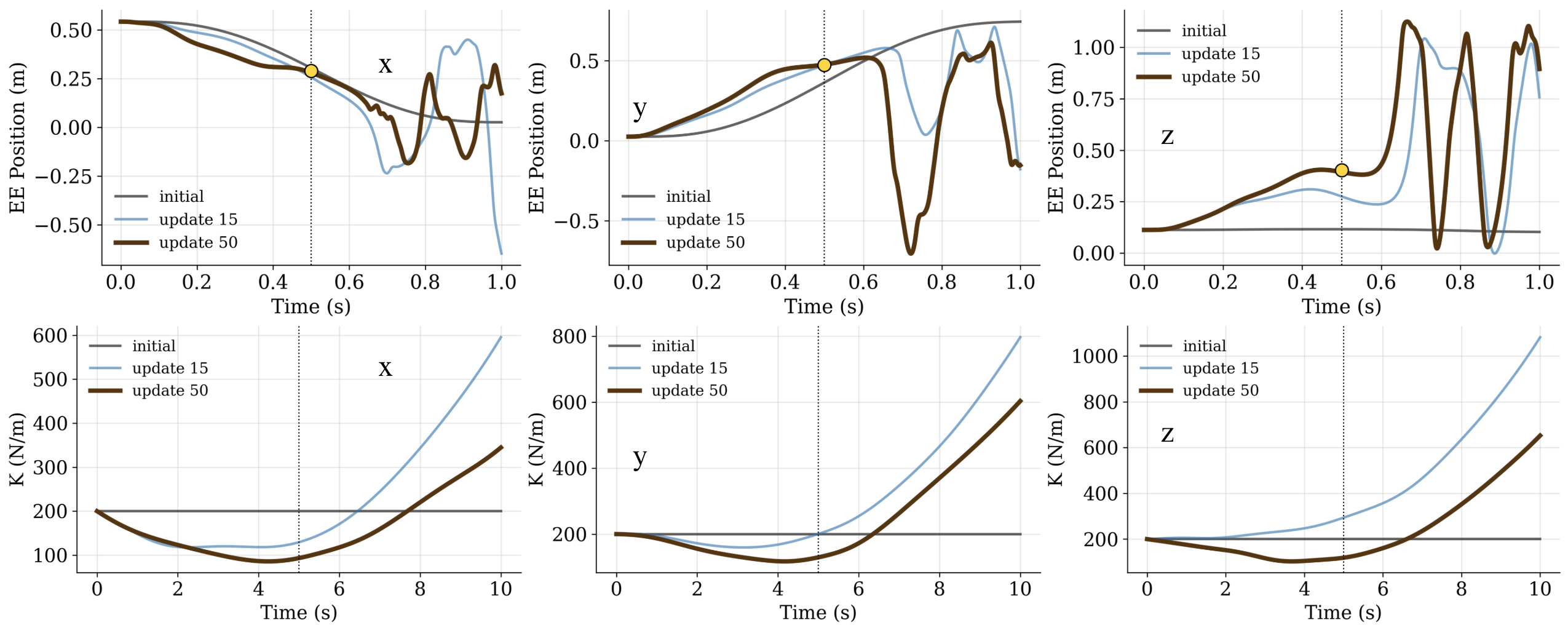}
    \caption{\footnotesize When C-GMS is applied only until the via-point (i.e., stability constraints are disabled afterward), the policy continues to optimize task cost but exhibits unstable behavior beyond the via-point, including oscillations and unbounded gain growth.}
    \label{fig:unsafe_res}
  \end{subfigure}
  \caption{VIC gain schedules and corresponding end-effector trajectories of the robot initially, after $15$ updates and after $50$ updates. The policy obtained after the $50^\mathrm{th}$ update was executed on hardware (cf. \S~\ref{subsec:conv}). Via-points are marked by circles.}
  \label{fig:results} 
  \vspace{-15pt}
\end{figure*}

In the initial phase (Fig.~\ref{fig:initial_hw}), the robot learns a smooth reaching motion from a starting pose $x_\mathrm{start} = [0.55,\, 0.00,\, 0.11]\,\mathrm{m}$ to a final handover location $x_\mathrm{goal} = [0.05,\, 0.72,\, 0.11]\,\mathrm{m}$, following a minimum-jerk trajectory in task space over a $10$ second horizon. This motion serves as the nominal demonstration and establishes a baseline for downstream optimization. To evaluate the system's adaptability, an obstacle is introduced along the nominal path, partially occluding the direct line between the start and goal. The robot must now adjust both its trajectory and impedance behavior to complete the handover safely. The adapted motion bends around the obstruction by passing through an intermediate region centered near $[0.30,\, 0.48,\, 0.40]\,\mathrm{m}$ (the obstacle’s coarse pose and was specified a priori, in practice they can be obtained automatically via open-vocabulary perception pipelines~\cite{liu2023grounding}). Fig.~\ref{fig:safe_hw} shows the physical configuration of the task space, including the obstacle location and handover geometry. We further evaluate generalization by executing five diverse free-space via-point policies under a consistent task setup, with quantitative results in Table~\ref{tab:hw_metrics}.

\subsection{Policy Representation}
We adopt the PI$^2$ as policy improvement framework (cf. \S\ref{subsec:prelim_pi2}) to jointly optimize the task-space trajectory and the time-varying Cartesian impedance gains. The policy is parameterized by a vector $\boldsymbol\theta$, consisting of DMP weights for the reference trajectory as well as slack parameters that indirectly control the stiffness and damping schedules as detailed in \S\ref{sec:method}.

The initial parameters for the trajectory are obtained by training the DMP to a minimum-jerk trajectory connecting $x_\mathrm{start}$ and $x_\mathrm{goal}$. The slack parameters are initialized such that the resulting stiffness matrix $\mathbf K(t)$ is constant and isotropic with magnitude $200\,\mathrm{N/m}$ along each axis. Damping is initialized at a constant value of $30\,\mathrm{Ns/m}$, also uniformly across axes. Complete parameter settings, including DMP bases, slack initializations, and cost weights are provided in Appendix B for reproducibility.

\subsection{Optimization Objective}
We formulated the cost function to primarily support via-point tracking. To achieve this, the objective function explicitly emphasizes tracking accuracy near the via-point, while also regularizing stiffness magnitude and motion smoothness. The total cost over a trajectory of length $T$ is:
$$
J = \sum_{t=1}^T \left[ 
\lambda_K\, \operatorname{tr}(\mathbf K_t) 
+ \lambda_{\text{acc}} \|\ddot{\mathbf x}_t\|^2 
+ w_{\text{via}}(t)\, \|\mathbf x_t - \mathbf x_{\text{ref},\,t}\|^2 
\right],
$$
\begin{itemize}
    \item $\operatorname{tr}(\mathbf K_t)$ penalizes the magnitude of the task-space stiffness matrix at time $t$, encouraging compliant behavior.
    \item $\|\ddot{\mathbf x}_t\|^2$ regularizes acceleration, promoting smooth motion.
    \item The tracking error term $\|\mathbf x_t - \mathbf x_{\text{ref},t}\|^2$ is scaled by a time-varying weight $w_{\text{via}}(t)$, which intensifies the penalty near the via-point.
\end{itemize}
The weighting function is given by $w_{\text{via}}(t) = w_0 + \gamma \cdot g(t)$ where $g(t)$ is a Gaussian kernel centered at the via-point time $\hat t$.

Learning is performed over $50$ policy updates, each with $12$ sampled rollouts in MuJoCo, a physics engine for simulation using the FR$3$ model. At every iteration, the PI$^2$ update rule is applied to the parameter vector $\boldsymbol \theta$. All sampling is constrained to a certified manifold that satisfies the time-varying stability condition (cf. Eq.~\eqref{eq:KB}). This ensures that all explored policies yield stabilizing behavior under the VIC dynamics in Eqs.~\eqref{eq:os-plant}-\eqref{eq:xdd_cmd}.

\begin{figure*}[h]
    \begin{subfigure}{0.24\linewidth}
    \centering
    \includegraphics[width=\textwidth]{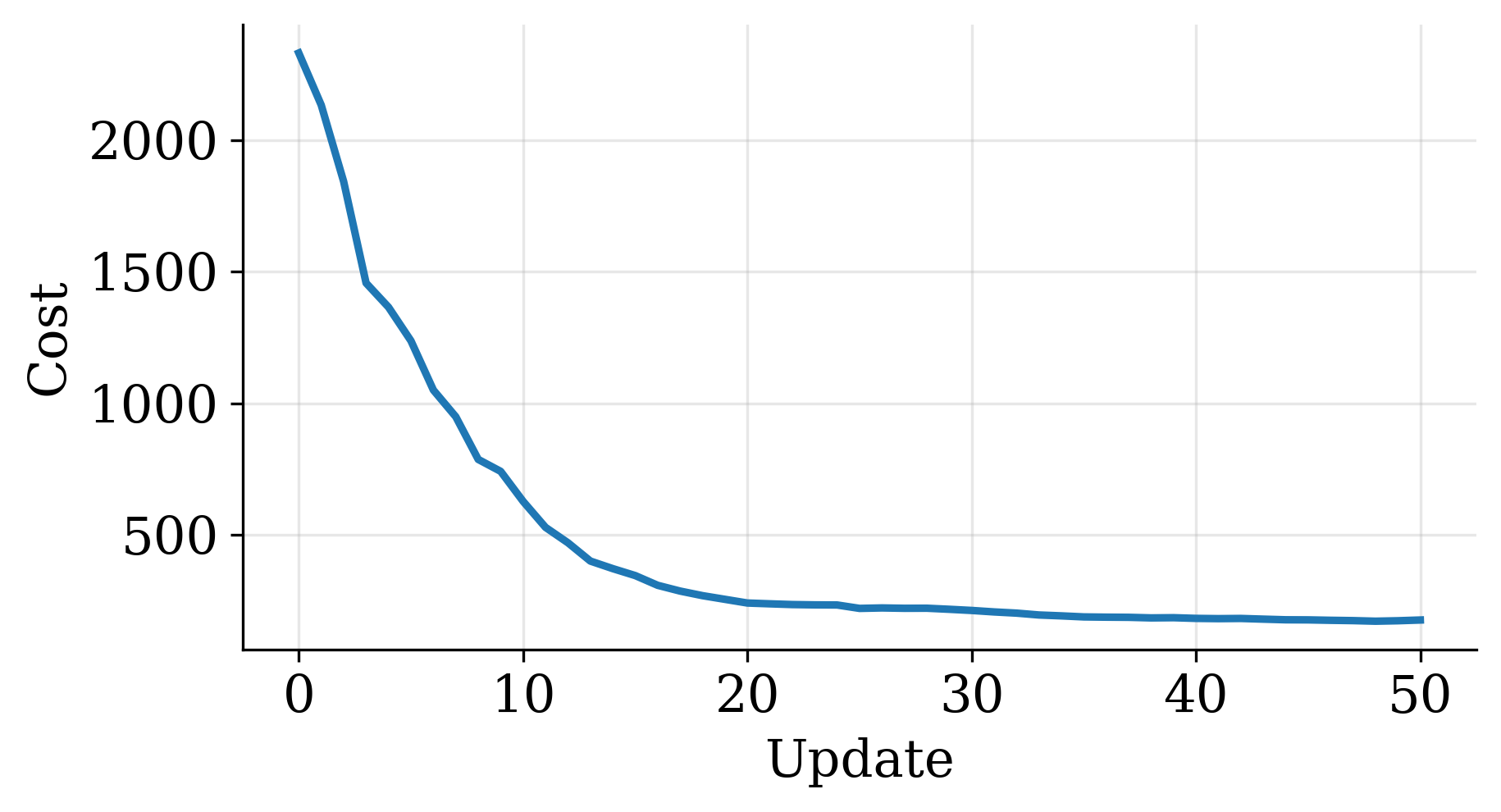}
    \caption{\footnotesize Learning Curve (C-GMS)}
    \label{fig:lc_safe}
  \end{subfigure}
  \hfill
  \begin{subfigure}{0.24\linewidth}
    \centering
    \includegraphics[width=\textwidth]{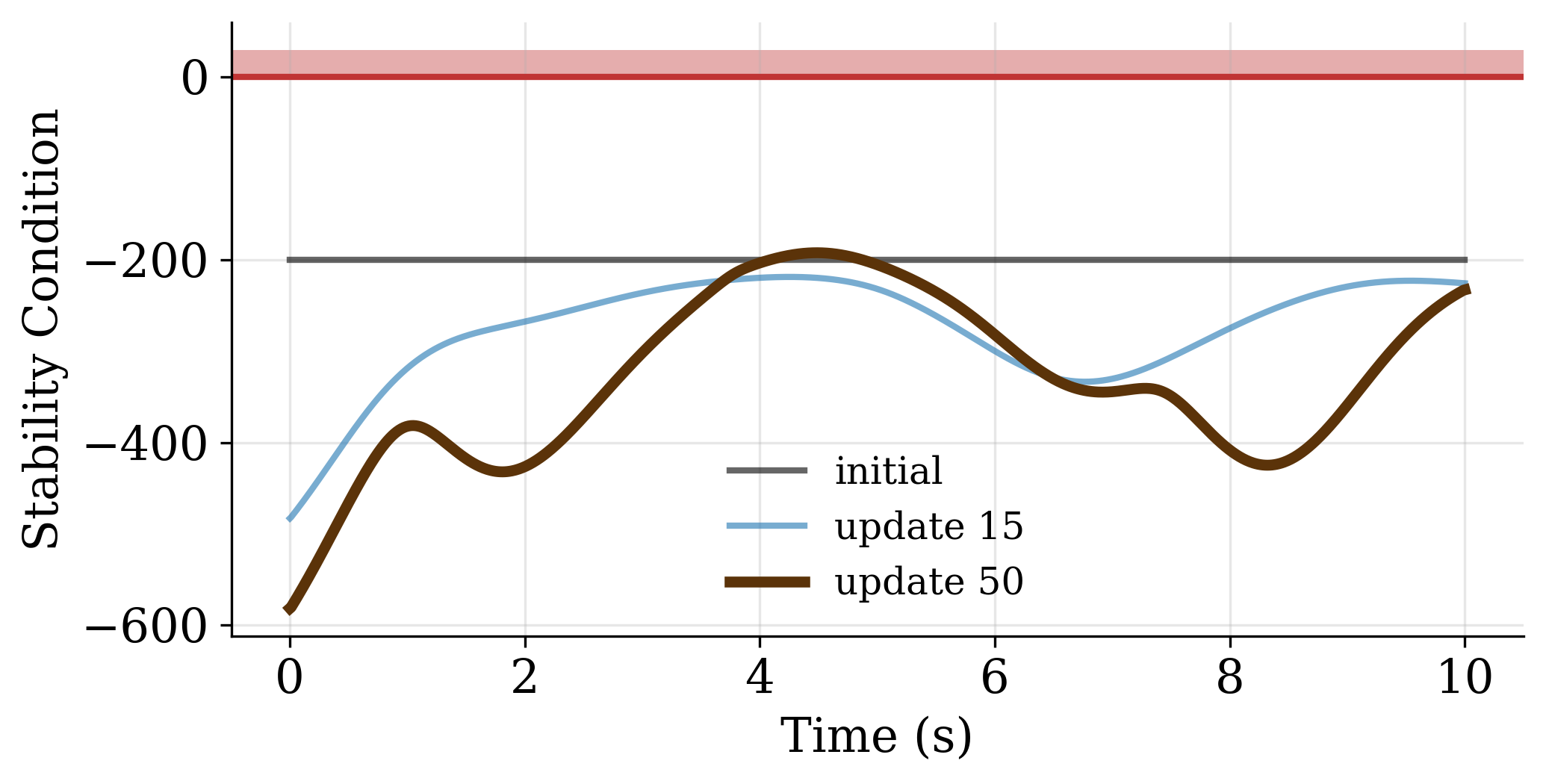}
    \caption{\footnotesize Stability Condition (C-GMS)}
    \label{fig:sc_safe}
  \end{subfigure}
  \hfill
  \begin{subfigure}{0.24\linewidth}
    \centering
    \includegraphics[width=\textwidth]{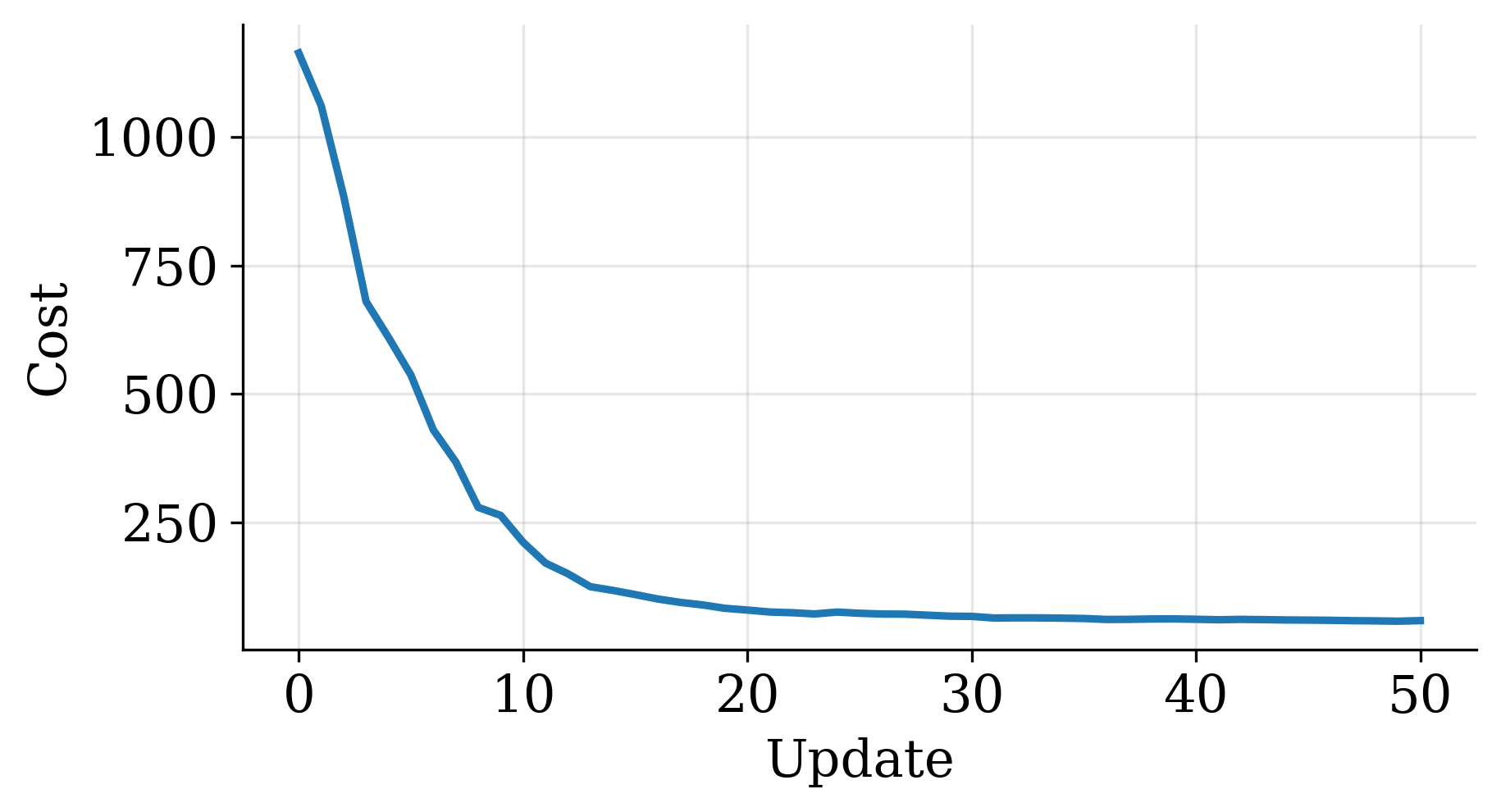}
    \caption{\footnotesize Learning Curve (no C-GMS)}
    \label{fig:lc_unsafe}
  \end{subfigure}
  \hfill
  \begin{subfigure}{0.24\linewidth}
    \centering
    \includegraphics[width=\textwidth]{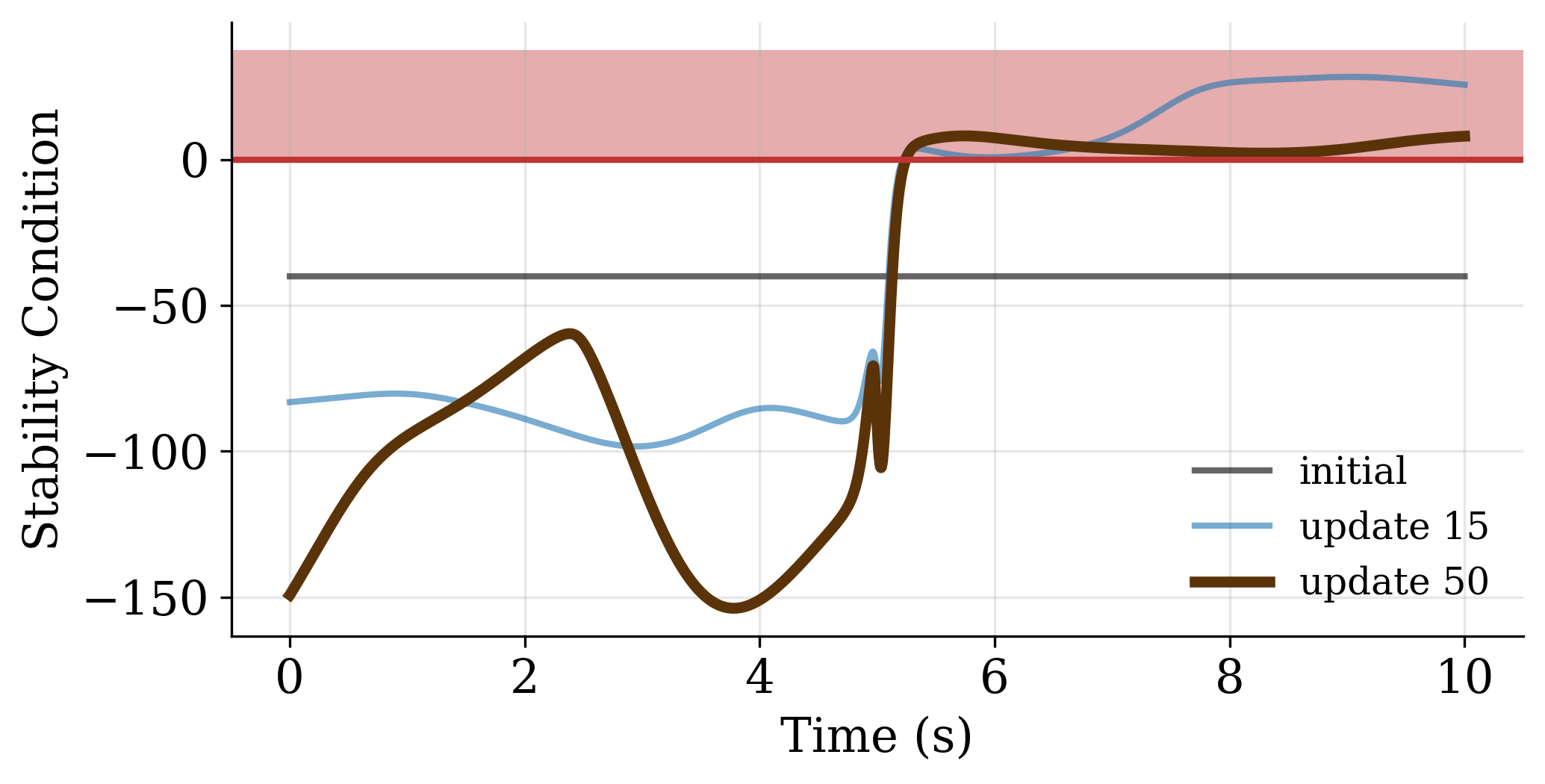}
    \caption{\footnotesize Stability Condition (no C-GMS)}
    \label{fig:sc_unsafe}
  \end{subfigure}
  \caption{Learning curve and eigenvalue evolution for Eq.~\eqref{eq:KB}. Under C-GMS based sampling, the eigenvalues remain negative, ensuring that the impedance profile guarantees stable control. However, when sampling shifts to an unsafe region, the cost function may still converge (since the via-point is reached prior to C-GMS being disabled), but the eigenvalues become positive, potentially leading to severe instability in the end-effector trajectory.}
  \label{fig:lc_and_sc}
  \vspace{-15pt}
\end{figure*}

\subsection{Results and Analysis}
\begin{table}[h]
\centering
\begin{minipage}{\columnwidth}
\footnotesize
\vspace{1mm}
\centering
\scriptsize
\setlength{\tabcolsep}{4pt}
\begin{tabular}{lccccc}
\toprule
Scenario & RMSE $x$  & RMSE $y$  & RMSE $z$  & Sat. w/ Gov & Sat. w/o Gov \\
\midrule
S1 & $2\text{e}{-2}$ & $53\text{e}{-4}$ & $54\text{e}{-4}$ & \xmark & \xmark \\
S2 & $38\text{e}{-4}$ & $35\text{e}{-4}$ & $76\text{e}{-4}$ & \xmark & \cmark \\
S3 & $11\text{e}{-3}$ & $22\text{e}{-3}$ & $3\text{e}{-2}$ & \xmark & \xmark \\
S4 & $27\text{e}{-4}$ & $19\text{e}{-4}$ & $28\text{e}{-4}$ & \xmark & \cmark \\
S5 & $23\text{e}{-3}$ & $51\text{e}{-4}$ & $1\text{e}{-2}$ & \xmark & \cmark \\
\bottomrule
\end{tabular}
\caption{\footnotesize Hardware metrics across five unique scenarios [Start-Via-End]:  \\
S1: $[0.30, 0.00, 0.47]-[0.42, 0.30, 0.34]-[0.54, 0.43, 0.47]$,     \\
S2: $[0.37, -0.34, 0.03]-[0.62, 0.00, 0.32]-[0.45, 0.27, 0.06]$,    \\
S3: $[0.40, 0.00, 0.15]-[0.32, 0.50, 0.42]-[0.00, 0.40, 0.10]$,     \\
S4: $[0.20, 0.17, 0.43]-[0.34, 0.20, 0.36]-[0.48, 0.34, 0.0.43]$,   \\
S5: $[0.58, -0.35, 0.18]-[0.31, 0.00, 0.43]-[0.00, 0.56, 0.05]$.     \\
RMSE-$x,\,y,\,z$ denotes the end-effector tracking error (in m) in task space. The final two columns indicate whether torque saturation was observed during execution, with and without the actuator-limit governor. Note that with governor, saturation was not reached for any segment.\\
\textbf{Note:} To evaluate the efficacy of \S~\ref{subsec:cags}, we virtually reduced FR$3$'s default torque limits ($[87,\,87,\,87,\,87,\,12,\,12,\,12]$\,Nm) by half, emulating deployment on lower-capacity hardware. The proposed governor maintains certification without exceeding these reduced limits.
}
\label{tab:hw_metrics}
\vspace{-25pt}
\end{minipage}
\end{table}
Figure~\ref{fig:hardware} demonstrates the task execution on real hardware. With certification (Fig.~\ref{fig:safe_hw}), the learned policy follows a stable trajectory through the via-point and completes the handover safely. Without certification (Fig.~\ref{fig:unsafe_hw}), the learned policy violates the Lyapunov condition and produces unstable trajectories eading to unsafe behaviors, including collisions with the environment/human. Figure~\ref{fig:results} compares the evolution of end-effector trajectories and stiffness profiles under the two regimes. With C-GMS (Fig.~\ref{fig:safe_res}), trajectories converge smoothly and stiffness rises near the via-point before relaxing, aligning naturally with task demands. Without certification (Fig.~\ref{fig:unsafe_res}), trajectories become oscillatory and gains diverge, despite reductions in cost. Figure~\ref{fig:lc_and_sc} summarizes optimization across both cases. Under C-GMS (Figs.~\ref{fig:lc_safe}, \ref{fig:sc_safe}), the cost decreases steadily while the minimum eigenvalue of Eq.~\eqref{eq:KB} remains strictly negative, certifying stability at every iteration. Without certification (Figs.~\ref{fig:lc_unsafe}, \ref{fig:sc_unsafe}), even though the cost converges, the eigenvalue crosses into the positive domain, indicating loss of guaranteed stability. This explains the unsafe executions in Fig.~\ref{fig:unsafe_res}. Table~\ref{tab:hw_metrics} summarizes performance of trained policy on hardware across five unique segments apart from the aforementioned collaborative task.



Together, these results show that while unconstrained learning can converge in terms of cost, it may yield unstable and unsafe policies. By embedding stability constraints directly into the sampling process, not only policies are guaranteed to be stable, but are also physically realizable on hardware, enabling the safe discovery of variable impedance policies.

\section{CONCLUSION AND LIMITATIONS}
\label{sec:dis}

We have introduced Certified Gaussian-Manifold Sampling (C-GMS), a novel framework that demonstrates how stable and optimal policy optimization can be achieved for variable impedance control by constraining the reinforcement learning exploration to a certified manifold. By leveraging the analytically verifiable  stability criterion, our approach guarantees that every policy rollout is Lyapunov stable and physically realizable. The results highlight a critical distinction: even with identical cost shaping and initialization, an unconstrained optimizer continues to reduce task cost but can produce unsafe gain schedules that lead to erratic and unstable robot execution. In contrast, C-GMS ensures the policy converges smoothly while preserving formal safety guarantees at each update. Our experiments further validate that C-GMS produces compliant trajectories that respect task constraints, showcasing its practical viability for safe autonomous interaction. The integration of certificate-aware actuator-limit governor presents a robust foundation for a physically realizable and safe learning system. 


A primary limitation of our current framework is its reliance on the stability criterion \cite{KBVIC}, which is formulated for free-space dynamics and does not account for external contact. This prevents its direct application to contact-rich tasks, a critical domain for compliant robotics. Additionally, the current cost function focuses solely on via-point tracking and lacks terms for orientation control, which is necessary for more realistic manipulation tasks.
Despite these limitations, our work opens several promising avenues for future research. We plan to extend this analysis to explore broader task families, including contact-rich and orientation-sensitive interactions. Future work will also explore how this governor can be extended to dynamically adapt to varying payload conditions.

\section*{APPENDIX}

\subsection{Robustness to Modeling Error}
\label{app:robustness}

During execution with OSID and wrench feedforward, the tracking error
\(\tilde{\mathbf{x}}=\mathbf{x}-\mathbf{x}_d\) evolves as
\begin{equation}
\mathbf{H}\,\ddot{\tilde{\mathbf{x}}}(t)+\mathbf{D}(t)\,\dot{\tilde{\mathbf{x}}}(t)+\mathbf{K}(t)\,\tilde{\mathbf{x}}(t)
= \mathbf{u_\mathrm{res}}(t),
\label{eq:A-dyn}
\end{equation}
where the residual input \(\mathbf{u_\mathrm{res}}(t)\) collects plant–model mismatch and feedforward/sensing imperfections and is bounded:
\(\|\mathbf{u_\mathrm{res}}(t)\|\le\bar{u}<\infty\).
Gains are synthesized by C-GMS from slacks with strict margins
\(\varepsilon_D,\varepsilon_K>0\) so that for all \(t\),
\begin{align}
\alpha\,\mathbf{H}-\mathbf{D}(t) &\preccurlyeq -\,\varepsilon_D\,\mathbf{I}, 
\label{eq:A-A}\\
\dot{\mathbf{K}}(t)+\alpha\,\dot{\mathbf{D}}(t)-2\alpha\,\mathbf{K}(t) 
&\preccurlyeq -\,\varepsilon_K\,\mathbf{I}.
\label{eq:A-C}
\end{align}
C-GMS guarantees \(\mathbf{K}(t)=\mathbf{Q}^\top\!(t)\mathbf{Q}(t)\succ0\) and \(\mathbf{D},\mathbf{K}\) are continuous.
On any horizon \([0,\,T]\), continuity implies uniform bounds
\[
h_{\min} \mathbf{I}\preccurlyeq \mathbf{H}\preccurlyeq h_{\max}\mathbf{I},\qquad
0<\underline{k}\mathbf{I}\preccurlyeq \mathbf{K}(t)\preccurlyeq \bar{k}\mathbf{I}.
\]
\textbf{Assumption:} \(\mathbf{D}(t)\) is uniformly bounded,
\(\|\mathbf{D}(t)\|\le \bar d<\infty\) (true if \(\mathbf{D}\) is continuous and its generating slacks are bounded).

Consider the standard energy
\begin{equation}
V(t)=\tfrac12\,\dot{\tilde{\mathbf{x}}}^{\!\top}\mathbf{H}\,\dot{\tilde{\mathbf{x}}}
+\tfrac12\,\tilde{\mathbf{x}}^{\!\top}\mathbf{K}(t)\,\tilde{\mathbf{x}}.
\label{eq:A-V}
\end{equation}
Because \(\mathbf{H}\succ0\) and \(\mathbf{K}(t)\succ0\), \(V\) is positive definite; there exist \(m_1,m_2>0\) such that
\begin{equation}
m_1\|z\|^2 \le V(t) \le m_2\|z\|^2,
\qquad
z:=\begin{bmatrix}\dot{\tilde{\mathbf{x}}}\\ \tilde{\mathbf{x}}\end{bmatrix}.
\label{eq:A-V-bounds}
\end{equation}
Differentiating \eqref{eq:A-V} along \eqref{eq:A-dyn} and using
\(\tilde{\mathbf{x}}^{\!\top}\mathbf{K}\dot{\tilde{\mathbf{x}}}
=\dot{\tilde{\mathbf{x}}}^{\!\top}\mathbf{K}\tilde{\mathbf{x}}\) gives
\begin{align}
\dot{V}
&= -\,\dot{\tilde{\mathbf{x}}}^{\!\top}\mathbf{D}\,\dot{\tilde{\mathbf{x}}}
+ \tfrac12\,\tilde{\mathbf{x}}^{\!\top}\dot{\mathbf{K}}\,\tilde{\mathbf{x}}
+ \dot{\tilde{\mathbf{x}}}^{\!\top}\mathbf{u_\mathrm{res}}.
\label{eq:A-Vdot-raw}
\end{align}
From \eqref{eq:A-A}, \(\mathbf{D}\succeq \alpha\mathbf{H}+\varepsilon_D\mathbf{I}\), hence
\begin{equation}
-\,\dot{\tilde{\mathbf{x}}}^{\!\top}\mathbf{D}\,\dot{\tilde{\mathbf{x}}}
\le -\,\alpha\,\dot{\tilde{\mathbf{x}}}^{\!\top}\mathbf{H}\,\dot{\tilde{\mathbf{x}}}
- \varepsilon_D\,\|\dot{\tilde{\mathbf{x}}}\|^2.
\label{eq:A-D-term}
\end{equation}
From \eqref{eq:A-C},
\begin{equation}
\tfrac12\,\tilde{\mathbf{x}}^{\!\top}\dot{\mathbf{K}}\,\tilde{\mathbf{x}}
\le \alpha\,\tilde{\mathbf{x}}^{\!\top}\mathbf{K}\,\tilde{\mathbf{x}}
- \tfrac{\alpha}{2}\,\tilde{\mathbf{x}}^{\!\top}\dot{\mathbf{D}}\,\tilde{\mathbf{x}}
- \tfrac{\varepsilon_K}{2}\,\|\tilde{\mathbf{x}}\|^2.
\label{eq:A-Kdot-ineq}
\end{equation}
Using
\(\tfrac{\alpha}{2}\,\tilde{\mathbf{x}}^{\!\top}\dot{\mathbf{D}}\,\tilde{\mathbf{x}}
=\tfrac{d}{dt}\big(\tfrac{\alpha}{2}\tilde{\mathbf{x}}^{\!\top}\mathbf{D}\tilde{\mathbf{x}}\big)
-\alpha\,\tilde{\mathbf{x}}^{\!\top}\mathbf{D}\dot{\tilde{\mathbf{x}}}\),
combine \eqref{eq:A-Vdot-raw}-\eqref{eq:A-Kdot-ineq} to obtain
\begin{align}
\dot{V} 
\le
&-\,\alpha\,\dot{\tilde{\mathbf{x}}}^{\!\top}\mathbf{H}\,\dot{\tilde{\mathbf{x}}}
- \varepsilon_D\,\|\dot{\tilde{\mathbf{x}}}\|^2
+ \alpha\,\tilde{\mathbf{x}}^{\!\top}\mathbf{K}\,\tilde{\mathbf{x}}
- \tfrac{d}{dt}\!\Big(\tfrac{\alpha}{2}\tilde{\mathbf{x}}^{\!\top}\mathbf{D}\tilde{\mathbf{x}}\Big)
\nonumber\\
&\hspace{12mm}
+ \alpha\,\tilde{\mathbf{x}}^{\!\top}\mathbf{D}\,\dot{\tilde{\mathbf{x}}}
- \tfrac{\varepsilon_K}{2}\,\|\tilde{\mathbf{x}}\|^2
+ \dot{\tilde{\mathbf{x}}}^{\!\top}\mathbf{u_\mathrm{res}}.
\label{eq:A-pre-aug}
\end{align}
Define the augmented storage
\(\displaystyle \mathcal{V}:=V+\tfrac{\alpha}{2}\,\tilde{\mathbf{x}}^{\!\top}\mathbf{D}\tilde{\mathbf{x}}\).
Then
\begin{align}
\dot{\mathcal{V}}
\le
&-\,\alpha\,\dot{\tilde{\mathbf{x}}}^{\!\top}\mathbf{H}\,\dot{\tilde{\mathbf{x}}}
- \varepsilon_D\,\|\dot{\tilde{\mathbf{x}}}\|^2
- \tfrac{\varepsilon_K}{2}\,\|\tilde{\mathbf{x}}\|^2
+ \alpha\,\tilde{\mathbf{x}}^{\!\top}\mathbf{K}\,\tilde{\mathbf{x}} \\
&+ \alpha\,\tilde{\mathbf{x}}^{\!\top}\mathbf{D}\,\dot{\tilde{\mathbf{x}}}
+ \dot{\tilde{\mathbf{x}}}^{\!\top}\mathbf{u_\mathrm{res}}.
\label{eq:A-Vaug}
\end{align}
Use \(\tilde{\mathbf{x}}^{\!\top}\mathbf{K}\tilde{\mathbf{x}}\le \bar{k}\,\|\tilde{\mathbf{x}}\|^2\),
\(\dot{\tilde{\mathbf{x}}}^{\!\top}\mathbf{H}\dot{\tilde{\mathbf{x}}}\ge h_{\min}\|\dot{\tilde{\mathbf{x}}}\|^2\),
\(\|\mathbf{D}(t)\|\le \bar d\),
Young’s inequality \(\alpha\,\tilde{\mathbf{x}}^{\!\top}\mathbf{D}\dot{\tilde{\mathbf{x}}}
\le \frac{\gamma}{2}\|\dot{\tilde{\mathbf{x}}}\|^2+\frac{\alpha^2\bar d^2}{2\gamma}\|\tilde{\mathbf{x}}\|^2\),
and \(\dot{\tilde{\mathbf{x}}}^{\!\top}\mathbf{u_\mathrm{res}}\le \eta\|\dot{\tilde{\mathbf{x}}}\|^2+\frac{1}{4\eta}\|\mathbf{u_\mathrm{res}}\|^2\)
for any \(\gamma\in(0,\varepsilon_D)\) and \(\eta\in(0,\varepsilon_D-\gamma)\). Then
\begin{align}
\dot{\mathcal{V}}
&\le
-\big(\alpha h_{\min}+\varepsilon_D-\gamma-\eta\big)\,\|\dot{\tilde{\mathbf{x}}}\|^2 \nonumber \\
&-\Big(\tfrac{\varepsilon_K}{2}-\alpha\bar{k}-\tfrac{\alpha^2 \bar d^2}{2\gamma}\Big)\,\|\tilde{\mathbf{x}}\|^2 
+ \tfrac{1}{4\eta}\,\|\mathbf{u_\mathrm{res}}(t)\|^2.
\label{eq:A-core-ineq}
\end{align}
Choose margins so that \(\varepsilon_K>2\alpha\bar{k}+\frac{\alpha^2 \bar d^2}{\gamma}\),
and fix any \(\gamma\in(0,\varepsilon_D)\), \(\eta\in(0,\varepsilon_D-\gamma)\).
Let
\[
c_1:=\min\!\Big\{\alpha h_{\min}+\varepsilon_D-\gamma-\eta,\ \tfrac{\varepsilon_K}{2}-\alpha\bar{k}-\tfrac{\alpha^2 \bar d^2}{2\gamma}\Big\}>0,
\]
\[
c_2:=\tfrac{1}{4\eta}>0.
\]
Then \eqref{eq:A-core-ineq} becomes
\begin{equation}
\dot{\mathcal{V}} \le -\,c_1\,\|z\|^2 + c_2\,\|\mathbf{u_\mathrm{res}}(t)\|^2.
\label{eq:A-final-ineq}
\end{equation}
Because \(\mathbf{K}\succ0\) and \(\mathbf{D}\succeq \alpha\mathbf{H}+\varepsilon_D\mathbf{I}\),
\(\mathcal{V}\) is positive definite and quadratically bounded with respect to \(z\).
Explicitly,
\begin{align}
&m_1'\|z\|^2 \le \mathcal{V}(t) \le m_2'\|z\|^2,  \nonumber \\
&m_1'=\tfrac12\min\{h_{\min},\,\underline{k}+\alpha\varepsilon_D\} \nonumber  \\
&m_2'=\tfrac12\max\{h_{\max},\,\bar{k}+\alpha\bar d\}.
\label{eq:A-Vaug-bounds}
\end{align}
By a comparison lemma applied to \eqref{eq:A-final-ineq}, for all \(t\ge t_0\),
\begin{equation}
\|z(t)\|^2 \le 
\exp\!\Big(-\tfrac{c_1}{m_2'}(t-t_0)\Big)\,\tfrac{\mathcal{V}(t_0)}{m_1'}
+\tfrac{m_2'}{m_1'}\,\tfrac{c_2}{c_1}\,\|\mathbf{u_\mathrm{res}}\|_\infty^2.
\label{eq:A-UUB}
\end{equation}
Thus, \((\tilde{\mathbf{x}},\dot{\tilde{\mathbf{x}}})\) is uniformly ultimately bounded (input-to-state practically stable), with ultimate radius \(O(\|\mathbf{u_\mathrm{res}}\|_\infty)\) that shrinks as the strict margins \((\varepsilon_D,\varepsilon_K)\) grow.
\hfill\(\square\)

\subsection{Hyperparameters and Reproducibility}
\label{subsec:appx_hyperparam}

Table~\ref{tab:hyperparams} lists all key hyperparameters used in the experiments. All values were held constant across trials and across all variants. No manual tuning was performed post-deployment. Code and configuration files are available at: \url{https://github.com/shr-eyas/safe-vic}

\begin{table}[h]
\centering
\caption{Key hyperparameters used in all experiments.}
\begin{tabular}{p{0.48\linewidth} p{0.42\linewidth}}
\toprule
\textbf{Parameter} & \textbf{Value} \\
\midrule
Time step $dt$                  & $0.001$\,s \\
Certificate scaling $\alpha$    & $0.05$ \\
Task matrix $H$                 & $\mathbf{I}_{3\times3}$ \\
RBF count (DMP)                 & $51$ \\
RBF count (slacks)              & $7$ \\
RBF intersection height (DMP)   & $0.95$ \\
RBF intersection height (slacks) & $0.7$ \\
RBF regularization          & $1\text{e}{-6}$ \\
PI$^2$ softmax sharpness $\beta$ & $20$ \\
Covariance decay (EMA)  & $0.98$ \\
Cost coefficient: $\lambda_K$ & $15\text{e}{-7}$ \\
Cost coefficient: $\lambda_{\textrm{acc}}$  & $1\text{e}{-3}$ \\
Cost coefficient: $w_0$  & $0.2$  \\
Cost coefficient: $\gamma$ & $5\text{e}{4}$ \\
Trajectory noise $\sigma_{\text{traj}}$ & $8.0$ \\
Stiffness noise $\sigma_K$ & $1.3$ \\
Damping noise $\sigma_D$ & $0.6$ \\
\bottomrule
\end{tabular}
\label{tab:hyperparams}
\end{table}

\bibliographystyle{ieeetr} 
\bibliography{references}

\end{document}